\documentclass[letterpaper]{article} % DO NOT CHANGE THIS
\usepackage{aaai25}  % DO NOT CHANGE THIS
\usepackage{times}  % DO NOT CHANGE THIS
\usepackage{helvet}  % DO NOT CHANGE THIS
\usepackage{courier}  % DO NOT CHANGE THIS
\usepackage[hyphens]{url}  % DO NOT CHANGE THIS
\usepackage{graphicx} % DO NOT CHANGE THIS
\usepackage{natbib}  % DO NOT CHANGE THIS AND DO NOT ADD ANY OPTIONS TO IT
\usepackage{caption} % DO NOT CHANGE THIS AND DO NOT ADD ANY OPTIONS TO IT
\usepackage{algorithm}
\usepackage{algorithmic}
\usepackage{multirow}
\usepackage{booktabs}
\usepackage{amssymb}
\usepackage{amsmath}

\usepackage{newfloat}
\usepackage{listings}
\DeclareCaptionStyle{ruled}{labelfont=normalfont,labelsep=colon,strut=off} % DO NOT CHANGE THIS
\floatstyle{ruled}
\newfloat{listing}{tb}{lst}{}
\floatname{listing}{Listing}
\title{Track the Answer: Extending TextVQA from Image to Video with Spatio-Temporal Clues}
\newcommand*\samethanks[1][\value{footnote}]{\footnotemark[#1]}
\author {
    Yan Zhang\textsuperscript{\rm 1,\rm 3}\equalcontrib,  Gangyan Zeng\textsuperscript{\rm 4}\equalcontrib, Huawen Shen\textsuperscript{\rm 1,\rm 3}, Daiqing Wu\textsuperscript{\rm 1,\rm 3}, Yu Zhou\textsuperscript{\rm 2}\thanks{Corresponding authors: Yu Zhou and Can Ma.}, Can Ma\textsuperscript{\rm 1,\rm 3}\samethanks
}

\affiliations{
    \textsuperscript{\rm 1}Institute of Information Engineering, Chinese Academy of Sciences, China\\
    \textsuperscript{\rm 2}VCIP \& TMCC \& DISSec, College of Computer Science, Nankai University, China \\
    \textsuperscript{\rm 3}School of Cyber Security, University of Chinese Academy of Sciences, China \\
    \textsuperscript{\rm 4}School of Cyber Science and Engineering, Nanjing University of Science and Technology, China
    
    \{zhangyan2022,shenhuawen,macan\}@iie.ac.cn, yzhou@nankai.edu.cn, gyzeng@njust.edu.cn
}

\begin{document}
\maketitle

\begin{abstract}
Video text-based visual question answering (Video TextVQA) is a practical task that aims to answer questions by jointly reasoning textual and visual information in a given video. Inspired by the development of TextVQA in image domain, existing Video TextVQA approaches leverage a language model (e.g. T5) to process text-rich multiple frames and generate answers auto-regressively. Nevertheless, the spatio-temporal relationships among visual entities (including scene text and objects) will be disrupted and models are susceptible to interference from unrelated information, resulting in irrational reasoning and inaccurate answering. To tackle these challenges, we propose the TEA (stands for ``\textbf{T}rack th\textbf{E} \textbf{A}nswer'') method that better extends the generative TextVQA framework from image to video. TEA recovers the spatio-temporal relationships in a complementary way and incorporates OCR-aware clues to enhance the quality of reasoning questions. Extensive experiments on several public Video TextVQA datasets validate the effectiveness and generalization of our framework. TEA outperforms existing TextVQA methods, video-language pretraining methods and video large language models by great margins. 
\end{abstract}


\begin{links}
    \link{Code}{https://github.com/zhangyan-ucas/TEA}
    
\end{links}

\section{Introduction}
Understanding scene text in images and videos plays an important role in interpreting the physical world, and also is the key to solving many practical applications, such as self-driving vehicles, automatic news reporting and scene text editing \cite{zeng2024textctrl,li2024first}. As a representative scene text understanding task \cite{zeng2024focus}, text-based visual question answering (TextVQA) has attracted much interest from the communities and witnessed tremendous progress \cite{singh2019towards,tap,SAL,zeng2023beyond,SETS}. A majority of work focuses on TextVQA in image domain (i.e. Image TextVQA), which aims at answering questions related to the text in a given image. Traditional methods \cite{m4c,ssbaseline} adopt a discriminative framework, which predicts answers from Optical Character Recognition (OCR) tokens \cite{qin2023towards} or vocabulary words. Benefiting from the powerful representation of pretrained language model T5 \cite{T5}, recent works \cite{latr,FITB} address this task in a generative way, thus significantly boosting the accuracy and generalizability.

\begin{figure}[t]
  \centering
  \includegraphics[width=0.435\textwidth]{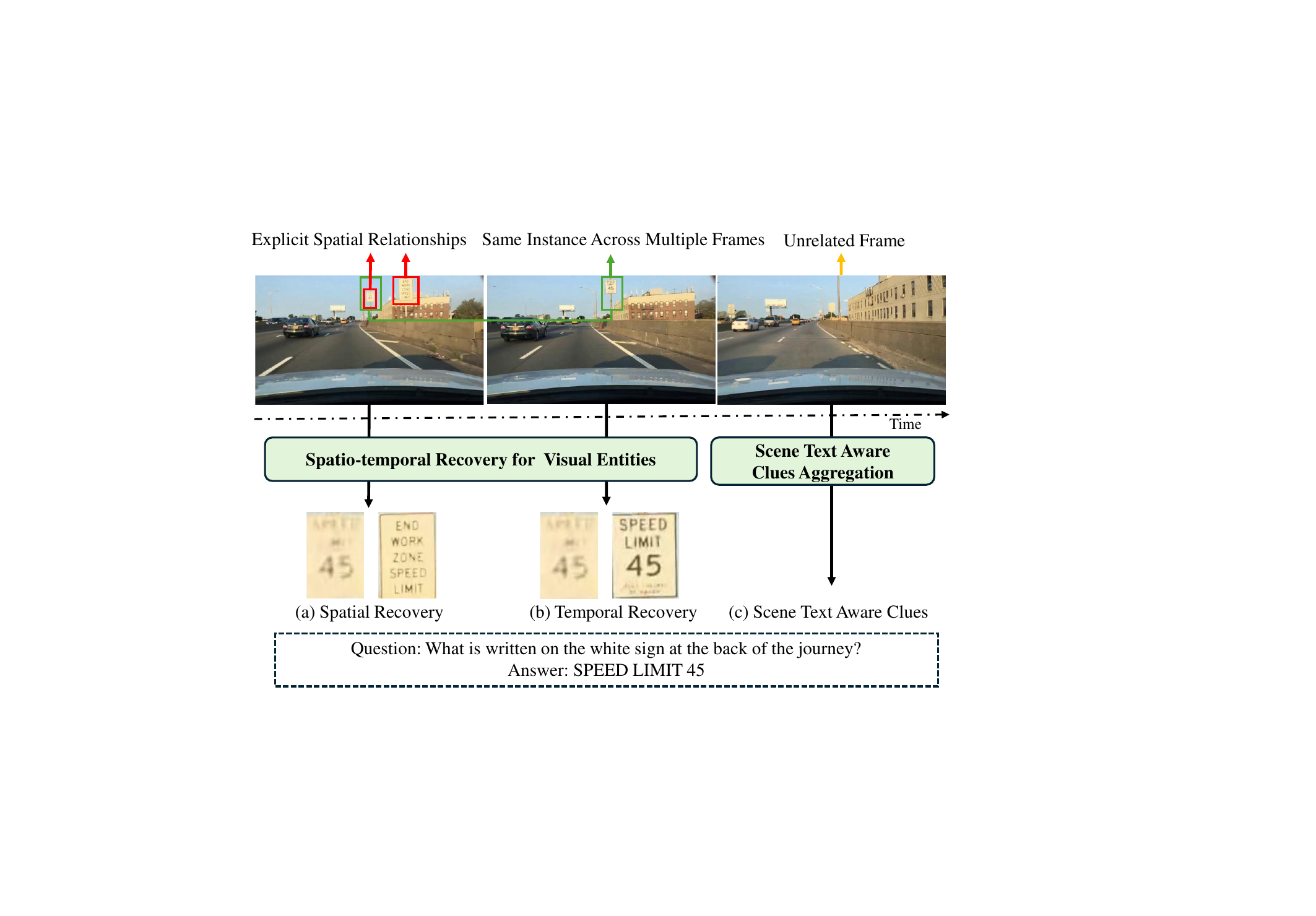}
  \vspace{-10pt}
  \caption{An example to illustrate the challenges in Video TextVQA and the overview of our proposed method. For this given question, the Video TextVQA model requires to first analyze the spatial relationships to locate the answer instance, and then capture the temporal correspondence to track and recognize the answer content. Besides, unrelated frames may interfere with model reasoning. 
  % To tackle these challenges, we propose to recover the spatio and temporal information respectively, and aggregate the scene text aware clues for reasonable guidance.
  }
  \vspace{-20pt}
\label{fig:figure1}
\end{figure}

In this context, a new task - Video TextVQA goes one step further, which requires reading and reasoning scene text throughout a video. Compared to Image TextVQA, this task is more challenging as the answer can appear anywhere in any frame of the video. To tackle it, an intuitive solution is to adapt the most competitive method for Image TextVQA to the video domain. Following it, T5-ViteVQA \cite{M4-ViteVQA} introduces five Transformer blocks to encode modality-specific features across video frames, and then feeds them into a T5-like Transformer encoder-decoder to infer the answer. 

Although the applicability of this vanilla pipeline has been demonstrated, there are still some limitations to consider. As illustrated in Fig. \ref{fig:figure1}, answering a Video TextVQA question often requires a comprehensive understanding of contextual information from multiple frames. On the one hand, the spatial relationships between entities within a frame (e.g., the two white signs in the first frame) should be correctly interpreted, and on the other hand, the instance correspondence between different frames (e.g., the sign ``SPEED LIMIT 45'' appearing in both the first and second frames) should be captured. Following general VideoQA's practice, current efforts tend to employ absolute-location embedding and temporal index to represent spatio and temporal information respectively. However, they ignore that Video TextVQA deviates from general VideoQA because a large amount of scene text will be involved. Considering scene text is relatively small and exhibits rich layout characteristics, this implicit learning method is hard to obtain satisfactory results. Besides, videos often involve a lot of information that is not related to answering the question (e.g., the unrelated third frame). This redundancy complicates the reasoning process, leading to challenges in analyzing intricate scenarios and filtering useless contents. Previous methods disperse the attention across the entire video and ignore the importance of question-guided reasoning cues, thus tending to predict inaccurate answers. 

To overcome the two aforementioned issues, we propose the \textbf{TEA} (stands for ``\textbf{T}rack th\textbf{E} \textbf{A}nswer'') method, which promotes the application of the T5-based generative framework to Video TextVQA in two aspects. 1) \textbf{Recover the spatio-temporal information.} We contend that previous generative Video TextVQA architectures organizing all visual entities into a 1-D sequence as input disrupt the spatio-temporal relationships that exist in the video. To this end, TEA recovers the spatio-temporal information in a complementary way. First, a temporal convolution module is utilized to fully exploit the continuity characteristics of visual entities in the temporal dimension, so as to capture the dynamic evolution of scene text across video frames. Then, we incorporate multi-granularity relative spatial relationships among scene text and explicitly encode them in the attention computation. In this manner, the composition of words inside each text instance and the spatial interaction between different entity instances can be easily attained. 2) \textbf{Aggregate the essential clues.} Understanding the intent of question and following a rational reasoning route are prerequisites for language model to generate correct answers. Following it, we design the scene text aware clues aggregation module, which progressively aggregates useful clues driven by the given question, suppressing the redundant message passing. Extensive experimental results on multiple public datasets underscore the effectiveness of our proposed framework. TEA significantly outperforms existing methods, including image and video TextVQA methods, video-language pretraining methods and video large language models (LLMs), marking a considerable advancement in the field of Video TextVQA.

The contributions of our work are threefold:
\begin{itemize}
\item 
Motivated by the development of Image TextVQA methods, we propose TEA, a generative Video TextVQA model that enables language model to understand 3-D video features and generate reasonable answers.

\item
To track the dynamic evolution of answers in the video, a temporal convolution module and an OCR-enhanced relative spatial module are developed to explicitly model spatio-temporal information among visual entities. To avoid interference from irrelevant information, a scene text aware clues aggregation module is utilized to extract the question-related reasoning clues.

\item Extensive experiments verify the effectiveness and superiority of our method TEA on several representative Video TextVQA datasets. 
\end{itemize}

\section{Related Work}
\subsection{Text-based Visual Question Answering}
\subsubsection{Image TextVQA. }
Image TextVQA task aims to answer questions by reading and understanding the textual contents in images.  Early Image TextVQA approaches \cite{m4c,sa-m4c,MM-GNN} predominantly predict answers from a fixed vocabulary list and OCR tokens \cite{pimnet,seed,tpsnet,instruction,cdistnet} present in the images. M4C \cite{m4c} leverages a Transformer-based multi-modal architecture and decodes answers auto-regressively with a dynamic pointer network. Built on the top of M4C, SA-M4C \cite{sa-m4c} incorporates the self-attention mask to enhance the spatial relationships among visual entities. Motivated by the success of visual-language multi-modal pre-training, TAP \cite{tap} designs Image TextVQA-specific pre-training tasks to align visual entities for scene text understanding. Recently, generative Image TextVQA approaches \cite{latr,FITB,SAL} have emerged which enhance the performance by leveraging powerful reasoning abilities and world knowledge from T5 \cite{T5}. For instance, \citet{latr} utilizes T5 to initialize the proposed LaTr model and address the question-answering problem in a text generation way.

\subsubsection{Video TextVQA. }  
Video TextVQA requires models to answer questions by jointly reasoning textual and visual information in a given video. Compared to Image TextVQA, Video TextVQA is more challenging as the answer can appear anywhere in any frame of the video. Additionally, the scene text in videos exhibits richer layout characteristics and more varied scales and forms. Motivated by the developments in generative Image TextVQA methods, T5-ViteVQA \cite{M4-ViteVQA} employs absolute-location embedding and temporal index to represent spatio-temporal among visual entities.  However, T5-ViteVQA fails to establish explicit spatio-temporal relationships among visual entities and efficiently process redundant visual entities in videos.

\begin{figure*}[t]
  \centering
  \includegraphics[width=0.8\textwidth]{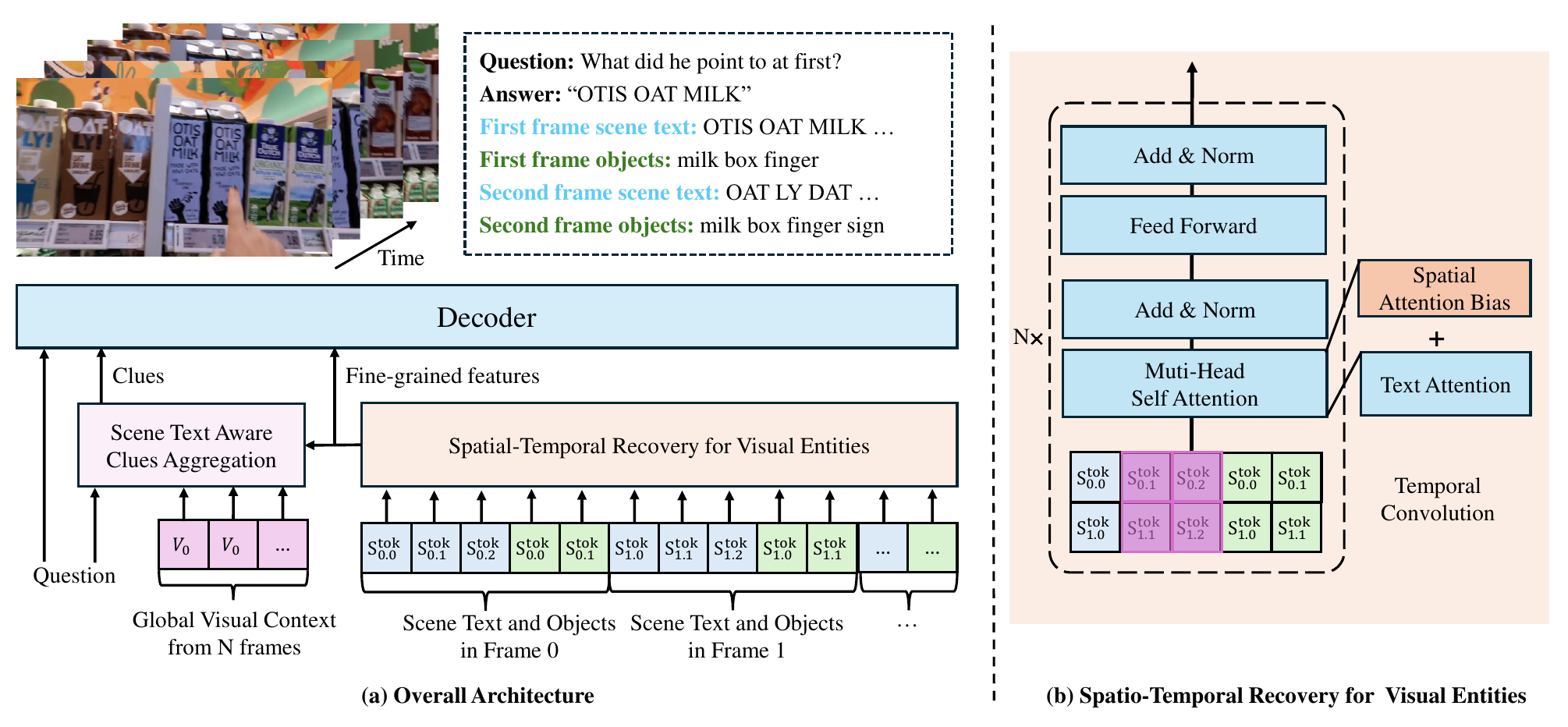}
  \vspace{-10pt}
  \caption{(a) Overall architecture of TEA. Blue and green token $S^{tok}_{i,j}$ represents the $j$th OCR and object token in $i$th frame, respectively. (b) Illustration of spatio-temporal recovery for visual entities. Temporal convolution and spatial attention bias represents the Temporal Convolution Module and OCR-enhanced relative spatial module, respectively. }
  \vspace{-10pt}
\label{fig:figure2}
\end{figure*}

\subsection{Video Question Answering}

Video question answering (VideoQA) is a crucial task in the visual-language domain, significantly enhancing AI systems' ability to interpret and interact with multimedia content. In recent years, VideoQA has attracted substantial interest from the research community, leading to various methodological advancements. Recent VideoQA methods can be categorized into several prominent approaches: 1) Memory Networks-based Methods: \citet{Gao2018MotionAppearanceCN} leverage memory networks to store and process long-term relationships across video frames.
2) Hierarchy Reasoning-based Methods: \citet{HierarchicalCR,VideoAC} hierarchically decompose questions and videos into sub-questions and video clips as well as derive the final answer based on the reasoning associated with each sub-question. 3) Transformer-based Methods: \citet{MIST} take advantage of the Transformer architecture in long-term relationships modeling to facilitate VideoQA research \cite{elysium}. 

Despite significant advancements in VideoQA, a common limitation persists: the focus predominantly remains on motions and events occurring in videos, disregarding textual information present in videos. Addressing this gap is crucial for developing AI systems capable of fully comprehending and interacting with a broader spectrum of visual content.

\section{Method}
\subsection{Overview Framework of TEA}
Fig. \ref{fig:figure2} illustrates the overall architecture of TEA, which primarily comprises visual external systems for extracting fine-grained visual entities and a T5-like Transformer encoder-decoder architecture for reasoning questions. For a given video, we uniformly sample it into multiple $T$ frames and utilize an OCR system \cite{TransVTSpotter,ABINET,hi-sam} to determine the location and content of scene text while concurrently employing an object detector \cite{faster-rcnn} to extract objects as auxiliary information to better understand the scene text.

Following Image TextVQA methods \cite{FITB}, we leverage a pre-trained language model \cite{T5} as the foundational model for the Video TextVQA task. To harness the reasoning capabilities and world knowledge of the foundational model, we convert the fine-grained features for Video TextVQA into a textual sequence to achieve input alignment. Specifically, a TEA-Baseline method is proposed, where we concatenate the contents of scene text and the labels of visual entities frame by frame into a sequence, as shown in Fig. \ref{fig:figure2}. Consequently, in TEA-Baseline, the video is represented as $ v \in \mathbb{R}^{T (M+N) \times d } $, where $T$ indicates the number of frames, $d$ denotes the feature dimension of visual entities, $M$ and $N$ indicate the number of scene text and objects within each video frame. However, simply feeding a sequence of Video TextVQA features into the foundational model leads to two issues: the disruption of spatio-temporal relationships among visual entities and the inability to effectively process long video sequences that include redundant fine-grained features. Therefore, TEA proposes two significant modules in the following section to recover the spatio-temporal relationships and generate scene text-aware clues to assist in generating answers.

\subsection{Spatio-Temporal Recovery for  Visual Entities}
In this section, we propose two sub-modules to separately recover temporal and spatial relationships. Specifically, a convolution operator is introduced to exploit the continuity characteristics of visual entities in the temporal dimension. Additionally, an OCR-enhanced spatial bias is incorporated into the attention mechanism to emphasize relative spatial relationships.

\subsubsection{Temporal Convolution Module}
Typically, T5-based Image TextVQA methods do not account for temporal modeling. To address this limitation, we design a temporal convolution module to reason about the temporal relationships between the same instance across different video frames. This design adheres to two crucial criteria: ensuring interaction between the same instance in different frames, and avoiding modifying the T5 representation space to maintain its reasoning abilities and world knowledge. Based on these principles, as shown in Fig. \ref{fig:figure2}(b), this sub-module employs a disentangled strategy to separately recover temporal and spatial relationships among visual entities and inserts a temporal convolution module between consecutive attention layers.

To be concrete, given the fine-grained features of a video ($ v \in \mathbb{R}^{T (M+N) \times d} $) with two dimensions of T(M+N) and d, we convert it into $ v \in \mathbb{R}^{T \times (M+N) \times d } $ with three dimensions of T, (M+N), and d.. A temporal convolution module is proposed to recover temporal relationships for the same instance across different video frames as follows: 

\begin{equation}
v' = v + \textit{Conv2D}(v \times W_{down}) \times W_{up},
\end{equation}
where $ v' \in \mathbb{R}^{T \times (M+N) \times d} $ is the temporal recovered video feature and \textit{Conv2D} denotes 2D-convolution. $W_{down}$ and $W_{up}$ indicate the projections for reducing and increasing dimensions. Specifically, a down-projection and up-projection strategy is applied before and after the temporal convolution. To capture temporal relationships for all visual entities across frames, the kernel size of the temporal convolution is designed with consideration of both the length of visual entities and the temporal length of videos. Furthermore, it is worth noting that benefiting from the strategy for disentangled spatio-temporal modeling, the computational complexity is also significantly reduced. \footnote{Joint spatio-temporal modeling complexity $O(v) = T^2(M+N)^2$ \textit{vs.} Disentangled spatio-temporal modeling complexity $O(v) = T(M+N)^2$}

\subsubsection{OCR-enhanced Relative Spatial Module}
\label{section3.2}
In the realm of text-rich understanding analysis, relative spatial relationships for visual entities are predominantly central \cite{BROS}, particularly for multi-granularity spatial relationships involving scene text \cite{hi-sam}. For instance, a representative question in Video TextVQA, such as \textit{``What is written on the white sign at the back of the journey?" }  in Fig. \ref{fig:figure1}, requires the model to locate relative spatial relationships between the last white sign and the text on it. Meanwhile recognizing ``SPEED", ``LIMIT", ``45" belong to the same paragraph-level bounding box allows for the establishment of associated semantic information. Therefore, we propose a novel OCR-enhanced relative spatial module that explicitly recovers the multi-granularity spatial relationships during attention calculations.

Following previous Image TextVQA methods built on the top of T5 encoder, we apply for multi-head self-attention mechanism to interact with multi-modal features and introduce an attention bias indicating relative spatial relationships, defined as:
\begin{equation}
\label{eq:e1}
%\[
e_{ij} = \frac{ Q_{i} * K_{j}^T }{\sqrt{d_z}} +a_{ij},
%\]
\end{equation} 
where $ e_{ij} \in \mathbb{R}^{head}$  represents the multi-head attention,  $a_{ij}$ indicates the relative spatial bias, $d_z$ represents the hidden dimension. $Q$ and $ K$ denote multi-head self-attention query and key for visual entities $i,j$, respectively. Note that in Image TextVQA methods $a_{ij}$ indicates 1-D relative relationships, while in TEA it indicates 2-D relative relationships.

Generally, we leverage visual external systems \cite{faster-rcnn, hi-sam} to obtain spatial locations for scene text and visual objects.  For formal description, denoting the normalized bounding box for each visual entity as $(x_{tl}, y_{tl},x_{br}, y_{br})$, inspired by \cite{BROS}, we compute relative spatial bias $a_{ij} \in \mathbb{R}^{head} $ with sinusoidal functions \cite{attention}: 

\begin{equation}
\label{eq:e2}
a_{ij} = [f^{sin}(x_i - x_j ); f^{cos}(y_i - y_j )].
\end{equation}

To enhance the semantic coherence in answering questions, we incorporate multi-granularity spatial relationships for scene text within the relative positional embedding. Specifically, we provide bounding boxes at the word, line, and paragraph levels for calculating the relative positional embeddings. Consequently, the term $a_{ij}$ in the equation \ref{eq:e2} can be expressed as follows:

% \begin{equation}
%        a_{ij} = 
%        \left \{
%         \begin{align*}
%             & a_{ij}^{word} + a_{ij}^{line} + a_{ij}^{para} 
%             \text{if i, j} \in \text{scene text} .\\ 
%             & a_{ij}^{word} \quad{   } \quad{   } \quad{   } \quad{} \quad{   } \quad{   } \quad{   }\text{otherwise.}\\
%         \end{align*}
%         \right
% \end{equation}

\begin{equation}
    a_{ij} = 
    \begin{cases}
        a_{ij}^{word} + a_{ij}^{line} + a_{ij}^{para} & \text{if } (i, j) \in \text{scene text,} \\ 
        a_{ij}^{word} & \text{otherwise.}
    \end{cases}
\end{equation}

\subsection{Scene Text Aware Clues Aggregation}
\begin{figure}
  \centering
  \includegraphics[width=0.33\textwidth]{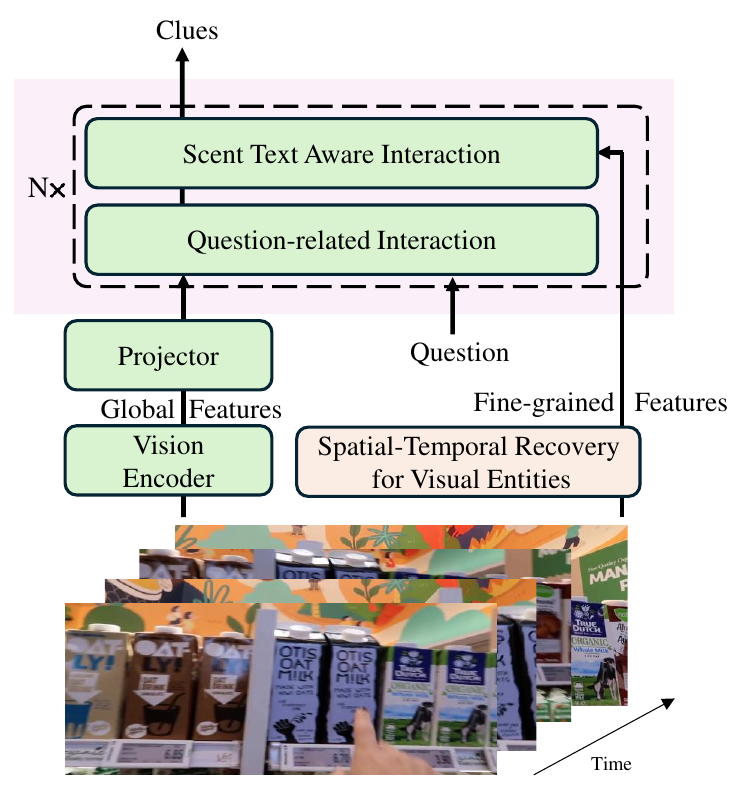}
  \vspace{-5pt}
  \caption{Illustration of scene text aware clues aggregation. }
  \vspace{-20pt}
    \label{fig:figure3}
\end{figure}

From the previous section, we can derive spatio-temporal features for visual entities. However, directly feeding these features into the decoder causes difficulties in answering certain questions, as it tends to focus too heavily on redundant fine-grained visual entities. Therefore, we propose a module that generates scene text-aware clues for the decoder to enhance the quality of reasoning within questions. In our design, we adhere to three practically crucial criteria: (1) simple and concise clues that indicate global context, (2) relevance to the question, and (3) awareness of fine-grained information, specifically related to scene text. This module is divided into two steps, as illustrated in Fig. \ref{fig:figure3}.

\begin{table*}[t]
\centering
\small
\begin{tabular}{ccccccc}
\toprule
\multirow{3}{*}{Method}           & \multicolumn{4}{c}{M4-ViteVQA \textit{Task1Split1} }                                          & \multicolumn{2}{c}{RoadTextVQA} \\ \cmidrule(l){2-5}  \cmidrule(l){6-7}
                                  & \multicolumn{2}{c}{Validation}         & \multicolumn{2}{c}{Test}                             & \multicolumn{2}{c}{Validation}          \\ \cmidrule(l){2-3} \cmidrule(l){4-5} \cmidrule(l){6-7}
                                  & Acc.(\%)       & ANLS(\%)           & Acc.(\%)       & ANLS(\%)                                 & Acc.(\%)        & ANLS(\%)            \\ \midrule
\multicolumn{1}{c|}{M4C \cite{m4c}}          & 18.66          & 24.20          & 17.91          & \multicolumn{1}{c|}{23.80}          & 28.92           & 32.27          \\ \midrule
\multicolumn{1}{c|}{JuskAsk \cite{Juskask}}      & 10.81          & 15.40          & 10.05          & \multicolumn{1}{c|}{14.10}          & --            & --           \\
\multicolumn{1}{c|}{All-in-one-B \cite{allinone}} & 11.47          & 15.30          & 10.87          & \multicolumn{1}{c|}{14.80}          & --            & --           \\
\multicolumn{1}{c|}{GIT \cite{GIT}}          & --           & --           & --           & \multicolumn{1}{c|}{--}           & 29.58           & 35.23          \\
\multicolumn{1}{c|}{SINGULARITY \cite{Singularity}}  & --           & --           & --           & \multicolumn{1}{c|}{--}           & 24.62           & 30.79          \\ \midrule
\multicolumn{1}{c|}{Video-LLaVA{*} \cite{videollava}}   & 15.82          & 17.77          & 15.43          & \multicolumn{1}{c|}{17.15}          & 30.82           & 40.92          \\
\multicolumn{1}{c|}{VideoLLaMA2{*} \cite{videollama2}}  & 20.04          & 21.73          & 20.76          & \multicolumn{1}{c|}{23.55}          & 25.11           & 36.53          \\ \midrule
\multicolumn{1}{c|}{T5-ViteVQA \cite{M4-ViteVQA}}   & 23.17          & 30.10          & 22.17          & \multicolumn{1}{c|}{29.10}          & --            & --           \\
\multicolumn{1}{c|}{TEA-Baseline} & 26.79          & 35.16          & 26.14          & \multicolumn{1}{c|}{35.14}          & 39.20           & 45.36          \\
\multicolumn{1}{c|}{TEA-Base}     & 34.45          & 42.91          & 31.70          & \multicolumn{1}{c|}{40.24}          & 44.43           & 51.69          \\
\multicolumn{1}{c|}{TEA-Large}    & \textbf{37.49} & \textbf{46.38} & \textbf{34.78} & \multicolumn{1}{c|}{\textbf{43.71}} & \textbf{48.14}  & \textbf{54.85} \\ \bottomrule
\end{tabular}
\caption{Accuracy and ANLS comparison with the latest methods on the M4-ViteVQA \textit{Task1Split1} and RoadTextVQA. The best results are highlighted in bold. {*} indicates the reproduced results using the open-source code.}
\label{table1}
\vspace{-15pt}
\end{table*}

\subsubsection{Question-related Global Context}
In the field of image recognition analysis, a single [CLS] token, positioned at the beginning of the visual encoder, represents the global context of an entire image. Therefore, we utilize the [CLS] token from the CLIP \cite{CLIP} vision encoder to capture the global context for visual semantics in each video frame. To guide these global context with Video TextVQA questions, we adopt a self-attention that integrates global features with question features. 

In practice, to generalize different sampling strategies, we design a projector to align multiple frames [CLS] token features into a fixed length. A self attention is proposed to extract global features relevant to the question. Specifically, we connect the video global context feature with the corresponding question to form a sequence. After fusion in self-attention, we extract the question-aware video global context as clues: 

% \begin{equation}
% \left\{
%         \begin{align*}
%             &X = [Global; Ques],\\
%             &X = SelfAttn(X),\\
%             &Clues = X[:\textit{len}(Global)],
%         \end{align*}
%         \right
% \end{equation}      

\begin{equation}
\left\{
\begin{aligned}
    &X = [Global; Ques], \\
    &X = SelfAttn(X), \\
    &Clues = X[:\textit{len}(Global)],
\end{aligned}
\right.
\end{equation}
where $[;]$ represents the concatenation of two features, and \textit{len} denotes the length of features.

\subsubsection{Scene Text Aware Interaction}
Based on question-related interaction, we obtain global features relevant to questions. To supplement the fine-grained details that might be overlooked by the CLIP visual encoder, we integrate them into global contexts with the cross-attention. Specifically, we employ scene text as the fine-grained feature to emphasize its association with the Video TextVQA questions:

\begin{equation}
X = CrossAttn(Q=X; K,V = visual \text{ } entities),\\
\label{eq:eq6}
\end{equation}
where $Q,K,V$ represent the query, key, and value in cross attention, respectively. Derived from the $N$-layer self-cross attention module, scene text-aware clues are then connected with spatio-temporal visual entities and fed into the decoder to generate answers.

\section{Experiments}
\subsection{Experiment Settings}
\subsubsection{Datasets and Evaluation Metrics}
Our method is rigorously evaluated using two public Video TextVQA datasets: M4-ViteVQA \cite{M4-ViteVQA} and  RoadTextVQA \cite{roadtextvqa}. 
1) M4-ViteVQA includes 8,511 video clips with 24,123 question-answer pairs, where the video clips are primarily collected from nine YouTube categories. Specifically, M4-ViteVQA defines two tasks with three settings to evaluate Video TextVQA methods. 
% In Task 1, Video TextVQA methods are trained and tested across all nine categories of M4-ViteVQA. To address different requirements for model robustness, \cite{M4-ViteVQA} designates \textit{Task1Split1} for regular testing and \textit{Task1Split2} for generalization testing, where the content of videos within the same category varies significantly (e.g., different shopping venues and sports). In Task 2, Video TextVQA methods are trained using seven categories and tested on the remaining two categories.
2) RoadTextVQA consists of 3,222 driving videos collected from multiple countries, annotated with 10,500 questions, all based on text or road signs present in the driving videos. For these datasets, we utilize the VQA Accuracy (Acc.) metric and the average normalized Levenshtein similarity (ANLS) to evaluate model performance of TEA.

\subsubsection{Implementation Details}
In our experimental setup, following previous Image TextVQA methods, we utilize pre-trained language model as the foundational backbones for processing scene text and objects as well as reasoning answers. Based on the pre-trained T5 \cite{T5}, we build two models with different capacities, namely TEA-Base and TEA-Large, which have 12+12 and 24+24 encoder-decoder layers, respectively.
% For frame sampling, we uniformly sample 10 frames for each dataset and standardize video frame inputs to a resolution of 224×224. 

\subsection{Comparison with Existing Methods}

\begin{table*}[t]
\small
\centering
\begin{tabular}{ccccccccc}
\toprule
\multirow{3}{*}{Method}           & \multicolumn{4}{c}{M4-ViteVQA \textit{Task1Split2} }                                             & \multicolumn{4}{c}{M4-ViteVQA \textit{Task2}}                              \\ \cmidrule(l){2-5}  \cmidrule(l){6-9}
                                  & \multicolumn{2}{c}{Validation}         & \multicolumn{2}{c}{Test}                             & \multicolumn{2}{c}{Validation}         & \multicolumn{2}{c}{Test}        \\ \cmidrule(l){2-3}  \cmidrule(l){4-5} \cmidrule(l){6-7} \cmidrule(l){8-9}
                                  & Acc.(\%)       & ANLS(\%)           & Acc.(\%)       & ANLS(\%)                                & Acc.(\%)       & ANLS(\%)           & Acc.(\%)       & ANLS(\%)           \\ \midrule
\multicolumn{1}{c|}{M4C \cite{m4c}}          & 13.58          & 17.20          & 11.36          & \multicolumn{1}{c|}{16.60}          & 9.16           & 12.80          & 7.52           & 12.50          \\ \midrule
\multicolumn{1}{c|}{JuskAsk \cite{Juskask}}      & 7.16           & 10.00          & 5.47           & \multicolumn{1}{c|}{8.60}           & 4.86           & 6.70           & 3.60           & 6.70           \\
\multicolumn{1}{c|}{All-in-one-B \cite{allinone}} & 6.85           & 9.20           & 5.66           & \multicolumn{1}{c|}{7.80}           & 4.20           & 5.00           & 3.28           & 4.60           \\ \midrule
\multicolumn{1}{c|}{Video-LLaVA{*} \cite{videollava}}   & 13.14          & 14.29          & 11.19          & \multicolumn{1}{c|}{12.02}          & 10.89          & 13.23          & 9.38           & 11.80          \\
\multicolumn{1}{c|}{VideoLLaMA2{*} \cite{videollama2}}  & 18.30          & 19.63          & 18.33          & \multicolumn{1}{c|}{20.45}          & 19.68          & 23.62          & 16.54          & 21.80          \\ \midrule
\multicolumn{1}{c|}{T5-ViteVQA \cite{M4-ViteVQA}}   & 17.59          & 23.10          & 16.68          & \multicolumn{1}{c|}{23.80}          & 12.30          & 16.10          & 9.29           & 13.60          \\
\multicolumn{1}{c|}{TEA-Baseline} & 22.53          & 31.60          & 21.20          & \multicolumn{1}{c|}{30.44}          & 17.32          & 26.32          & 14.00          & 21.99          \\
\multicolumn{1}{c|}{TEA-Base}     & 26.66          & 36.61          & 26.29          & \multicolumn{1}{c|}{36.00}          & 20.73          & 28.18          & 17.28          & 26.03          \\
\multicolumn{1}{c|}{TEA-Large}    & \textbf{28.27} & \textbf{36.32} & \textbf{28.43} & \multicolumn{1}{c|}{\textbf{38.13}} & \textbf{22.83} & \textbf{30.21} & \textbf{18.83} & \textbf{28.90} \\ \bottomrule
\end{tabular}
\caption{Accuracy and ANLS comparison with the latest methods on the M4-ViteVQA \textit{Task1Split2} and \textit{Task2}. The best results are highlighted in bold. {*} indicates the reproduced results using the open-source code.}
\label{table2}
\vspace{-10pt}
\end{table*}

On the M4-ViteVQA \textit{Task1Split1} and RoadTextVQA benchmarks, we evaluate TEA and compare it with existing methods (see Tab. \ref{table1}). It can be observed that TEA-Large outperforms different types of existing methods.
(1) \textbf{Image TextVQA Methods.} M4C \cite{m4c} is unable to perform temporal reasoning in videos, achieving an accuracy of 17.91\% on M4-ViteVQA test set.  (2) \textbf{Video-language Pre-training Methods.} These methods \cite{Juskask,allinone,GIT,Singularity} lack the ability to read and comprehend scene text in videos, resulting in unsatisfactory performance on the Video TextVQA task. All-in-one-B achieves an accuracy of 10.87\% on the M4-ViteVQA test set, while GIT achieves 29.58\% in accuracy on the RoadTextVQA validation set. (3) \textbf{Video-LLMs}. 
Video-LLMs \cite{videollama2, videollava,vila} build upon LLMs and leverage a large-scale of video-text pairs pre-training data to align textual and video modalities. To adapt these methods for Video TextVQA, we
apply LoRA \cite{lora} to fine-tune on corresponding dataset.
A representative Video-LLMs method, VideoLLaMA2, achieves an accuracy of 20.04\% and an ANLS score of 21.73\% on the M4-ViteVQA validation set. As reported in the results, Video TextVQA methods are generally superior to Video-LLMs because Video-LLMs lack the capability to capture and understand massive small scene text in videos and do not incorporate architectures specifically designed for Video TextVQA. (4) \textbf{Video TextVQA Methods.} Compared with a pioneer Video TextVQA method T5-ViteVQA \cite{M4-ViteVQA}, TEA-Baseline shows a 3.97\% accuracy improvement on M4-ViteVQA test set. Although T5 \cite{T5} is our common foundation model, TEA-Baseline better aligns multimodal inputs by utilizing textual representation for each visual entity. Compared to TEA-Baseline, TEA-Base recovers the spatio-temporal relationships and provides instructional clues for generating answers, resulting in significant accuracy gains of 5.56\% on the M4-ViteVQA test set and 5.23\%  on the RoadTextVQA validation set. Finally, when upgrading the model capacity to a large size, TEA-Large reaches state-of-the-art performance, 34.78\% and 48.14\% on the M4-ViteVQA test set and RoadTextVQA validation set, respectively.

\begin{table}[]
\centering
\small
\begin{tabular}{cccccc}
\toprule
\multirow{2}{*}{\#} & \multirow{2}{*}{Temporal} & \multirow{2}{*}{Spatial} & \multirow{2}{*}{Clue}     & \multicolumn{2}{c}{M4-ViteVQA}  \\ \cmidrule{5-6} 
                    &                           &                          &                           & Acc.(\%)       & ANLS(\%)           \\ \midrule
a                   &                           &                          & \multicolumn{1}{c|}{}     & 26.79          & 35.16          \\
b                   & \checkmark                      &                          & \multicolumn{1}{c|}{}     & 28.56          & 37.51          \\
c                   &                           & \checkmark                     & \multicolumn{1}{c|}{}     & 31.91          & 40.93          \\
d                   & \checkmark                      & \checkmark                     & \multicolumn{1}{c|}{}     & 33.69          & 42.91          \\
e                   &                           &                          & \multicolumn{1}{c|}{\checkmark} & 29.22          & 38.06          \\
f                   & \checkmark                      & \checkmark                     & \multicolumn{1}{c|}{\checkmark} & \textbf{34.45} & \textbf{43.35} \\ \bottomrule
\end{tabular}

\caption{Overall ablation results on the M4-ViteVQA \textit{Task1Split1} validation set. Temporal, Spatial, Clue represents temporal convolution module, OCR-enhanced relative spatial module, and scene text aware clues aggregation, respectively.}
\vspace{-15pt}
\label{table3}
\end{table}

\subsection{Analysis of Generalization}
In this section, we investigate how TEA contributes to the Video TextVQA from a generalization perspective. Most VideoQA methods tend to overfit specific training sets, achieving high accuracy on these trained video categories while failing to predict answers in other categories. However, in practical applications, it is essential for these methods to generalize their abilities across different domains.
% Adapting to other tasks is a critical component for Video TextVQA methods.
Following the experimental settings established by \cite{M4-ViteVQA}, we define two task splits to validate generalization capabilities. As illustrated in Tab. \ref{table2}, our method achieves consistent improvements in accuracy and ANLS compared to both visual-language pre-training methods and Video-LLMs. We attribute it in two factors. 
With the incorporation of pretrained language model \cite{T5}, TEA exploits its world knowledge in scene text descriptions.  Additionally, our devised spatio-temporal recovery for visual entities module explicitly establishes spatio-temporal relationships among visual entities, which are shared cross various scenarios, including even in unseen video categories.

\subsection{Ablation Studies}
\subsubsection{Overall Results}
To empirically validate the contribution of key components in TEA, i.e., spatio-temporal recovery for visual entities and scene text aware clues aggregation, we conduct ablation studies on the M4-ViteVQA \textit{Task1Split1} validation set. The corresponding modules are individually added, and their impact on the model’s performance is measured. The overall comparative results are detailed in Tab. \ref{table3} and the specific hyperparameters for each module will be discussed in the following sections. Firstly, we denote TEA-Baseline, where multimodal features for scene text and objects across video frames are connected into a sequence and fed into the foundational model T5, This approach disrupts the spatial and temporal relationships between visual entities (\#a). When the temporal convolution module, which provides temporal relationships between the same instance across video frames, is added, there is a 1.77\% increase in accuracy (\#b). In contrast, the OCR-enhanced relative spatial module results in a more substantial improvement of 5.12\% in accuracy (\#c). Subsequently, jointly recovering the spatio-temporal relationships leads to an advancement of 6.90\% in accuracy and 7.75\% in ANLS (\#d). To demonstrate the ability of aggregating fine-grained features, we ablate the scene text aware clues aggregation module in two settings: without and with spatio-temporal features. Specifically, there is a 2.43\% accuracy increase in the former setting (\#e) and a 0.76\% accuracy increase in the latter (\#f).

\subsubsection{The Effect of OCR-enhanced Relative Spatial Module}
\begin{table}[]
\small
\centering
\begin{tabular}{c|cc}
\toprule
Multi-granularity & Acc.(\%)   & ANLS(\%)  \\ \midrule
                  & 31.06 & 40.02 \\
\checkmark              & \textbf{31.91} & \textbf{40.93} \\ \bottomrule
\end{tabular}
\caption{Ablation experiments about OCR-enhanced relative spatial module on M4-ViteVQA \textit{Task1Split1} validation dataset. 
% Multi-granularity denotes mulit-granularity relationships among scene text. 
}
\label{table4}
\vspace{-5pt}
\end{table}

The goal of the OCR-enhanced relative spatial module is to establish spatial relationships between visual entities, particularly for scene text. By providing word-level, line-level, and paragraph-level spatial relationships, the model is guided to recognize which scene text contains relevant semantic information. In Tab. \ref{table4}, replacing these multi-granularity relationships with only word-level ones will lead to a decrease of 0.85\% in accuracy.

\subsubsection{The Effect of Scene Text Aware Clues Aggregation}

\begin{table}[]
\centering
\small
\begin{tabular}{cc|cc}
\toprule
\# & Features         & Acc.(\%)       & ANLS(\%)           \\ \midrule
a  & --             & 26.79          & 35.16          \\
b  & Ques             & 27.40          & 36.47          \\
c  & Ques + OCR + OBJ & 28.46          & 37.21          \\
d  & Ques + OCR       & \textbf{29.22} & \textbf{38.06} \\ \bottomrule
\end{tabular}

\caption{Ablation experiments about scene text aware clues aggregation on M4-ViteVQA \textit{Task1Split1} validation dataset. 
% Ques, OCR, OBJ represent the features of the question, scene text, objects, respectively. 
}
\vspace{-10pt}
\label{table5}
\end{table}

% The scene text aware clues aggregation module aims to aggregate the essential clues and incorporate counterparts to assist TEA in generating high-quality answers.
To assess the effectiveness of this module, we explore the features that need to be aggregated in Tab. \ref{table5}. We can notice that compared to the initial version with only global visual semantic features (\#a), integrating the question brings a 0.61\% accuracy improvement (\#b). This validates the pivotal role of questions in solving the Video TextVQA task, while the perception of fine-grained features is still insufficient. For this reason, by further incorporating the scene text and object features, a significant increase of 1.06\% is observed (\#c). Furthermore, we find that composing fined-grained features with scene text alone leads to an additional 0.76\% improvement (\#d). This might suggest that visual objects aid in comprehending the relative location and related semantics of scene text, yet they are ineffective in facilitating clues aggregation.

\subsection{Qualitative Analysis}
\begin{figure}[t]
    \centering  
  \includegraphics[width=0.50\textwidth]{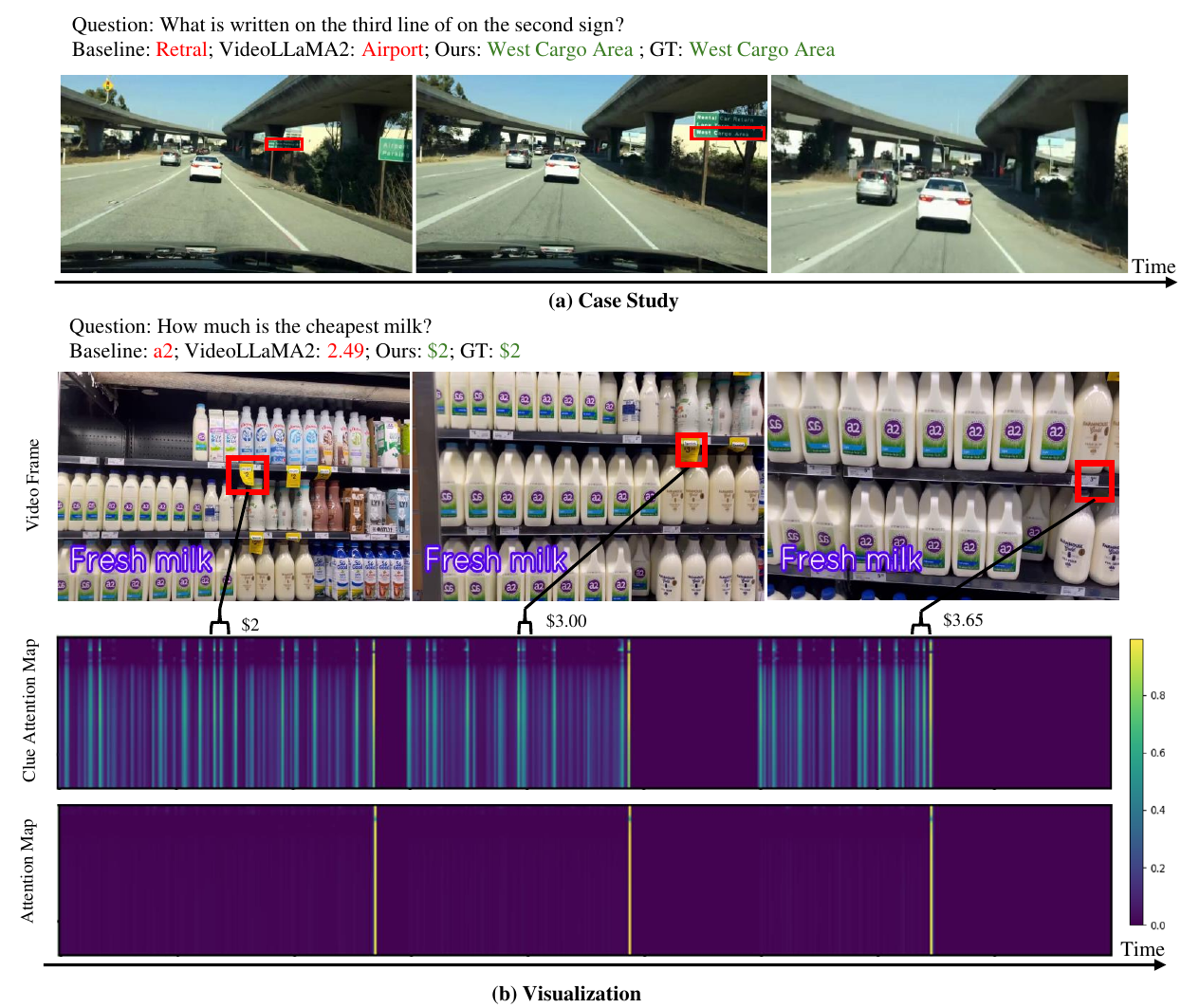}
  
  \caption{
  (a) A case study on M4-ViteVQA. The red boxes indicate the regions related to the question. (b) Visualization of cross-attention between questions and scene text across each video frame in the answer generation process. Red boxes in the raw video frames represent the locations of the corresponding highlighted scene text.
}
  \vspace{-10pt}
    \label{fig:figure4}
\end{figure}

\subsubsection{Case Study}
In Fig. \ref{fig:figure4}(a), a representative qualitative example is presented to provide an intuitive analysis of TEA in comparison with TEA-Baseline and VideoLLaMA2. We observe that TEA offers advantages in the following aspects: 1) Spatial-temporal relationships for scene text. TEA determines the relative spatial relationships between the first sign and the scene text associated with the second sign in the first frame. Additionally, since the scene text on the second sign in the first frame is not clearly visible, TEA establishes temporal relationships between the instances of the scene text ``West Cargo Area" to enhance its visibility. 2) Aggregation for the essential clues. TEA identifies the third frame is not related to the question and aggregates a clue to assist in answering. 

\subsubsection{Visualization}
To demonstrate the effectiveness of our scene text aware clues aggregation, we visualize the cross-attention involved in the answer generation process, as depicted in Fig. \ref{fig:figure4}(b). These visualizations provide valuable insights into how our module focuses on question-related visual entities within each video frame. The comparative analysis highlights a clear contrast between the refined attention maps (second row) and the initial attention maps (third row). Initially, the standard attention primarily highlights [SEP] token \cite{Clark2019WhatDB} and shows insufficient attention to question-related scene text. In contrast, the refined attention maps demonstrate a concentrated focus on question-related scene text.

\section{Conclusion}
In this paper, we present an initial investigation into extending Image TextVQA models to Video TextVQA methods through a simple yet highly effective approach, referred to as TEA. TEA aims to recover spatio-temporal relationships between scene texts across different video frames. We aggregate scene text aware clues to guide the model in identifying which frames are relevant to the questions. The capability of TEA to read and comprehend scene texts in videos also contributes to the advancement of understanding dynamic environments within the research community.

\section{Acknowledgments}
Supported by the National Natural Science Foundation of China (Grant NO 62376266 and 62406318), and by the Key Research Program of Frontier Sciences, CAS (Grant NO ZDBS-LY-7024).

\bibliography{aaai25}

\begin{thebibliography}{44}
\providecommand{\natexlab}[1]{#1}

\bibitem[{Biten et~al.(2021)Biten, Litman, Xie, Appalaraju, and Manmatha}]{latr}
Biten, A.~F.; Litman, R.; Xie, Y.; Appalaraju, S.; and Manmatha, R. 2021.
\newblock {LaTr: Layout-aware Transformer for Scene-text VQA}.
\newblock \emph{CVPR}, 16527--16537.

\bibitem[{Cheng et~al.(2024)Cheng, Leng, Zhang, Xin, Li, Chen, Zhu, Zhang, Luo, Zhao, and Bing}]{videollama2}
Cheng, Z.; Leng, S.; Zhang, H.; Xin, Y.; Li, X.; Chen, G.; Zhu, Y.; Zhang, W.; Luo, Z.; Zhao, D.; and Bing, L. 2024.
\newblock {VideoLLaMA2: Advancing Spatial-temporal Modeling and Audio Understanding in Video-LLMs}.
\newblock \emph{ArXiv}, abs/2406.07476.

\bibitem[{Clark et~al.(2019)Clark, Khandelwal, Levy, and Manning}]{Clark2019WhatDB}
Clark, K.; Khandelwal, U.; Levy, O.; and Manning, C.~D. 2019.
\newblock {What Does BERT Look At? An Analysis of BERT’s Attention}.
\newblock In \emph{ACL}.

\bibitem[{Du et~al.(2024)Du, Chen, Su, Jia, and Jiang}]{instruction}
Du, Y.; Chen, Z.; Su, Y.; Jia, C.; and Jiang, Y.-G. 2024.
\newblock Instruction-guided Scene Text Recognition.
\newblock \emph{arXiv preprint arXiv:2401.17851}.

\bibitem[{Fang et~al.(2023)Fang, Li, Li, Ma, and Hu}]{SAL}
Fang, C.; Li, J.; Li, L.; Ma, C.; and Hu, D. 2023.
\newblock {Separate and Locate: Rethink the Text in Text-based Visual Question Answering}.
\newblock \emph{ACM MM}, 4378--4388.

\bibitem[{Fang et~al.(2021)Fang, Xie, Wang, Mao, and Zhang}]{ABINET}
Fang, S.; Xie, H.; Wang, Y.; Mao, Z.; and Zhang, Y. 2021.
\newblock {Read Like Humans: Autonomous, Bidirectional and Iterative Language Modeling for Scene Text Recognition}.
\newblock \emph{CVPR}, 7094--7103.

\bibitem[{Gao et~al.(2020)Gao, Li, Wang, Shan, and Chen}]{MM-GNN}
Gao, D.; Li, K.; Wang, R.; Shan, S.; and Chen, X. 2020.
\newblock {Multi-modal Graph Neural Network for Joint Reasoning on Vision and Scene Text}.
\newblock In \emph{CVPR}, 12746--12756.

\bibitem[{Gao et~al.(2022)Gao, Zhou, Ji, Zhu, Yang, and Shou}]{MIST}
Gao, D.; Zhou, L.; Ji, L.; Zhu, L.; Yang, Y.; and Shou, M.~Z. 2022.
\newblock {MIST: Multi-modal Iterative Spatial-temporal Transformer for Long-form Video Question Answering}.
\newblock \emph{CVPR}, 14773--14783.

\bibitem[{Gao et~al.(2018)Gao, Ge, Chen, and Nevatia}]{Gao2018MotionAppearanceCN}
Gao, J.; Ge, R.; Chen, K.; and Nevatia, R. 2018.
\newblock {Motion-appearance Co-memory Networks for Video Question Answering}.
\newblock \emph{CVPR}, 6576--6585.

\bibitem[{Hong et~al.(2021)Hong, Kim, Ji, Hwang, Nam, and Park}]{BROS}
Hong, T.; Kim, D.; Ji, M.; Hwang, W.; Nam, D.; and Park, S. 2021.
\newblock {BROS: A Pre-trained Language Model Focusing on Text and Layout for Better Key Information Extraction from Documents}.
\newblock In \emph{AAAI}, 10767--10775.

\bibitem[{Hu et~al.(2021)Hu, Shen, Wallis, Allen-Zhu, Li, Wang, Wang, and Chen}]{lora}
Hu, E.~J.; Shen, Y.; Wallis, P.; Allen-Zhu, Z.; Li, Y.; Wang, S.; Wang, L.; and Chen, W. 2021.
\newblock {LoRA: Low-rank Adaptation of Large Language Models}.
\newblock In \emph{ICLR}.

\bibitem[{Hu et~al.(2019)Hu, Singh, Darrell, and Rohrbach}]{m4c}
Hu, R.; Singh, A.; Darrell, T.; and Rohrbach, M. 2019.
\newblock {Iterative Answer Prediction with Pointer-augmented Multimodal Transformers for TextVQA}.
\newblock \emph{CVPR}, 9989--9999.

\bibitem[{Kant et~al.(2020)Kant, Batra, Anderson, Schwing, Parikh, Lu, and Agrawal}]{sa-m4c}
Kant, Y.; Batra, D.; Anderson, P.; Schwing, A.; Parikh, D.; Lu, J.; and Agrawal, H. 2020.
\newblock {Spatially Aware Multimodal Transformers for TextVQA}.
\newblock In \emph{ECCV}, 715--732.

\bibitem[{Le et~al.(2020)Le, Le, Venkatesh, and Tran}]{HierarchicalCR}
Le, T.~M.; Le, V.; Venkatesh, S.; and Tran, T. 2020.
\newblock Hierarchical Conditional Relation Networks for Video Question Answering.
\newblock \emph{CVPR}, 9969--9978.

\bibitem[{Lei, Berg, and Bansal(2023)}]{Singularity}
Lei, J.; Berg, T.; and Bansal, M. 2023.
\newblock {Revealing Single Frame Bias for Video-and-language Learning}.
\newblock In \emph{ACL}, 487--507.

\bibitem[{Li et~al.(2024)Li, Shu, Zeng, Yang, and Zhou}]{li2024first}
Li, Z.; Shu, Y.; Zeng, W.; Yang, D.; and Zhou, Y. 2024.
\newblock {First Creating Backgrounds then Rendering Texts: A New Paradigm for Visual Text Blending}.
\newblock In \emph{ECAI}.

\bibitem[{Lin et~al.(2023)Lin, Zhu, Ye, Ning, Jin, and Yuan}]{videollava}
Lin, B.; Zhu, B.; Ye, Y.; Ning, M.; Jin, P.; and Yuan, L. 2023.
\newblock {Video-LLaVA: Learning United Visual Representation by Alignment before Projection}.
\newblock \emph{ArXiv}, abs/2311.10122.

\bibitem[{Lin et~al.(2024)Lin, Yin, Ping, Molchanov, Shoeybi, and Han}]{vila}
Lin, J.; Yin, H.; Ping, W.; Molchanov, P.; Shoeybi, M.; and Han, S. 2024.
\newblock {VILA: On Pre-training for Visual Language Models}.
\newblock In \emph{CVPR}, 26689--26699.

\bibitem[{Qiao et~al.(2021)Qiao, Zhou, Wei, Wang, Zhang, Jiang, Wang, and Wang}]{pimnet}
Qiao, Z.; Zhou, Y.; Wei, J.; Wang, W.; Zhang, Y.; Jiang, N.; Wang, H.; and Wang, W. 2021.
\newblock {PIMNet: A Parallel, Iterative and Mimicking Network for Scene Text Recognition}.
\newblock In \emph{ACM MM}, 2046--2055.

\bibitem[{Qiao et~al.(2020)Qiao, Zhou, Yang, Zhou, and Wang}]{seed}
Qiao, Z.; Zhou, Y.; Yang, D.; Zhou, Y.; and Wang, W. 2020.
\newblock {SEED: Semantics Enhanced Encoder-decoder Framework for Scene Text Recognition}.
\newblock In \emph{CVPR}, 13528--13537.

\bibitem[{Qin et~al.(2023)Qin, Lyu, Zhang, Zhou, Yao, Zhang, Lin, and Wang}]{qin2023towards}
Qin, X.; Lyu, P.; Zhang, C.; Zhou, Y.; Yao, K.; Zhang, P.; Lin, H.; and Wang, W. 2023.
\newblock {Towards Robust Real-time Scene Text Detection: From Semantic to Instance Representation Learning}.
\newblock In \emph{ACM MM}, 2025--2034.

\bibitem[{Radford et~al.(2021)Radford, Kim, Hallacy, Ramesh, Goh, Agarwal, Sastry, Askell, Mishkin, Clark, Krueger, and Sutskever}]{CLIP}
Radford, A.; Kim, J.~W.; Hallacy, C.; Ramesh, A.; Goh, G.; Agarwal, S.; Sastry, G.; Askell, A.; Mishkin, P.; Clark, J.; Krueger, G.; and Sutskever, I. 2021.
\newblock {Learning Transferable Visual Models from Natural Language Supervision}.
\newblock In \emph{ICML}.

\bibitem[{Raffel et~al.(2019)Raffel, Shazeer, Roberts, Lee, Narang, Matena, Zhou, Li, and Liu}]{T5}
Raffel, C.; Shazeer, N.~M.; Roberts, A.; Lee, K.; Narang, S.; Matena, M.; Zhou, Y.; Li, W.; and Liu, P.~J. 2019.
\newblock Exploring the Limits of Transfer Learning with a Unified Text-to-text Transformer.
\newblock \emph{JMLR}, 21: 140:1--140:67.

\bibitem[{Ren et~al.(2015)Ren, He, Girshick, and Sun}]{faster-rcnn}
Ren, S.; He, K.; Girshick, R.~B.; and Sun, J. 2015.
\newblock {Faster R-CNN: Towards Real-time Object Detection with Region Proposal Networks}.
\newblock \emph{TPAMI}, 39: 1137--1149.

\bibitem[{Singh et~al.(2019)Singh, Natarajan, Shah, Jiang, Chen, Batra, Parikh, and Rohrbach}]{singh2019towards}
Singh, A.; Natarajan, V.; Shah, M.; Jiang, Y.; Chen, X.; Batra, D.; Parikh, D.; and Rohrbach, M. 2019.
\newblock Towards VQA Models that Can Read.
\newblock In \emph{CVPR}, 8317--8326.

\bibitem[{Tom et~al.(2023)Tom, Mathew, Garcia, Karatzas, and Jawahar}]{roadtextvqa}
Tom, G.; Mathew, M.; Garcia, S.; Karatzas, D.; and Jawahar, C. 2023.
\newblock {Reading Between the Lanes: Text VideoQA on the Road}.
\newblock In \emph{ICDAR}, 137--154.

\bibitem[{Vaswani(2017)}]{attention}
Vaswani, A. 2017.
\newblock {Attention is All You Need}.
\newblock In \emph{NeurIPS}.

\bibitem[{Wang et~al.(2022{\natexlab{a}})Wang, Ge, Yan, Ge, Lin, Cai, Wu, Shan, Qie, and Shou}]{allinone}
Wang, A.; Ge, Y.; Yan, R.; Ge, Y.; Lin, X.; Cai, G.; Wu, J.; Shan, Y.; Qie, X.; and Shou, M.~Z. 2022{\natexlab{a}}.
\newblock {All in One: Exploring Unified Video-language Pre-training}.
\newblock \emph{CVPR}, 6598--6608.

\bibitem[{Wang et~al.(2025)Wang, Ye, Wang, Nie, and Huang}]{elysium}
Wang, H.; Ye, Y.; Wang, Y.; Nie, Y.; and Huang, C. 2025.
\newblock {Elysium: Exploring Object-level Perception in Videos via MLLM}.
\newblock In \emph{ECCV}, 166--185.

\bibitem[{Wang et~al.(2022{\natexlab{b}})Wang, Yang, Hu, Li, Lin, Gan, Liu, Liu, and Wang}]{GIT}
Wang, J.; Yang, Z.; Hu, X.; Li, L.; Lin, K.; Gan, Z.; Liu, Z.; Liu, C.; and Wang, L. 2022{\natexlab{b}}.
\newblock {GIT: A Generative Image-to-text Transformer for Vision and Language}.
\newblock \emph{TMLR}.

\bibitem[{Wang et~al.(2022{\natexlab{c}})Wang, Zhou, Lv, Wu, Zhao, Jiang, and Wang}]{tpsnet}
Wang, W.; Zhou, Y.; Lv, J.; Wu, D.; Zhao, G.; Jiang, N.; and Wang, W. 2022{\natexlab{c}}.
\newblock {TPSNet: Reverse Thinking of Thin Plate Splines for Arbitrary Shape Scene Text Representation}.
\newblock In \emph{ACM MM}, 5014--5025.

\bibitem[{Wu et~al.(2021)Wu, Cai, Zhang, Wang, Li, Li, Tang, and Zhou}]{TransVTSpotter}
Wu, W.; Cai, Y.; Zhang, D.; Wang, S.; Li, Z.; Li, J.; Tang, Y.; and Zhou, H. 2021.
\newblock {A Bilingual, OpenWorld Video Text Dataset and End-to-end Video Text Spotter with Transformer}.
\newblock In \emph{NeurIPS}.

\bibitem[{Xiao et~al.(2021)Xiao, Yao, Liu, Li, Ji, and seng Chua}]{VideoAC}
Xiao, J.; Yao, A.; Liu, Z.; Li, Y.; Ji, W.; and seng Chua, T. 2021.
\newblock {Video as Conditional Graph Hierarchy for Multi-granular Question Answering}.
\newblock In \emph{AAAI}, 2804--2812.

\bibitem[{Yang et~al.(2020)Yang, Miech, Sivic, Laptev, and Schmid}]{Juskask}
Yang, A.; Miech, A.; Sivic, J.; Laptev, I.; and Schmid, C. 2020.
\newblock {Just Ask: Learning to Answer Questions from Millions of Narrated Videos}.
\newblock \emph{ICCV}, 1666--1677.

\bibitem[{Yang et~al.(2021)Yang, Lu, Wang, Yin, Florencio, Wang, Zhang, Zhang, and Luo}]{tap}
Yang, Z.; Lu, Y.; Wang, J.; Yin, X.; Florencio, D.; Wang, L.; Zhang, C.; Zhang, L.; and Luo, J. 2021.
\newblock {TAP: Text-aware Pre-training for TextVQA and Text Caption}.
\newblock In \emph{CVPR}, 8751--8761.

\bibitem[{Ye et~al.(2024)Ye, Zhang, Liu, Liu, Yin, Liu, Du, and Tao}]{hi-sam}
Ye, M.; Zhang, J.; Liu, J.; Liu, C.; Yin, B.; Liu, C.; Du, B.; and Tao, D. 2024.
\newblock {Hi-SAM: Marrying Segment Anything Model for Hierarchical Text Segmentation}.
\newblock \emph{TPMAI}.

\bibitem[{Zeng et~al.(2024{\natexlab{a}})Zeng, Zhang, Wei, Yang, Zhang, Gao, Qin, and Zhou}]{zeng2024focus}
Zeng, G.; Zhang, Y.; Wei, J.; Yang, D.; Zhang, P.; Gao, Y.; Qin, X.; and Zhou, Y. 2024{\natexlab{a}}.
\newblock {Focus, Distinguish, and Prompt: Unleashing CLIP for Efficient and Flexible Scene Text Retrieval}.
\newblock In \emph{ACM MM}, 2525--2534.

\bibitem[{Zeng et~al.(2023{\natexlab{a}})Zeng, Zhang, ZHOU, Fang, Zhao, Wei, and Wang}]{FITB}
Zeng, G.; Zhang, Y.; ZHOU, Y.; Fang, B.; Zhao, G.; Wei, X.; and Wang, W. 2023{\natexlab{a}}.
\newblock {Filling in the Blank: Rationale-augmented Prompt Tuning for TextVQA}.
\newblock \emph{ACM MM}, 1261--1272.

\bibitem[{Zeng et~al.(2023{\natexlab{b}})Zeng, Zhang, Zhou, Yang, Jiang, Zhao, Wang, and Yin}]{zeng2023beyond}
Zeng, G.; Zhang, Y.; Zhou, Y.; Yang, X.; Jiang, N.; Zhao, G.; Wang, W.; and Yin, X.-C. 2023{\natexlab{b}}.
\newblock {Beyond OCR+VQA: Towards End-to-end Reading and Reasoning for Robust and Accurate TextVQA}.
\newblock \emph{PR}, 138: 109337.

\bibitem[{Zeng et~al.(2024{\natexlab{b}})Zeng, Shu, Li, Yang, and Zhou}]{zeng2024textctrl}
Zeng, W.; Shu, Y.; Li, Z.; Yang, D.; and Zhou, Y. 2024{\natexlab{b}}.
\newblock {TextCtrl: Diffusion-based Scene Text Editing with Prior Guidance Control}.
\newblock In \emph{NeurIPS}.

\bibitem[{Zhang et~al.(2024)Zhang, Zeng, Shen, Ma, and Zhou}]{SETS}
Zhang, Y.; Zeng, G.; Shen, H.; Ma, C.; and Zhou, Y. 2024.
\newblock Show Exemplars and Tell Me What You See: In-context Learning with Frozen Large Language Models for TextVQA.
\newblock In \emph{PRCV}, 231--245.

\bibitem[{Zhao et~al.(2022)Zhao, Li, Wang, Li, Zhou, Zhang, Xuyang, Yu, Yu, Li, Dai, and Zhou}]{M4-ViteVQA}
Zhao, M.; Li, B.; Wang, J.; Li, W.; Zhou, W.; Zhang, L.; Xuyang, S.; Yu, Z.; Yu, X.; Li, G.; Dai, A.; and Zhou, S. 2022.
\newblock {Towards Video Text Visual Question Answering: Benchmark and Baseline}.
\newblock In \emph{NeurIPS}, 35549--35562.

\bibitem[{Zheng et~al.(2024)Zheng, Chen, Fang, Xie, and Jiang}]{cdistnet}
Zheng, T.; Chen, Z.; Fang, S.; Xie, H.; and Jiang, Y.-G. 2024.
\newblock {CDistNet: Perceiving Multi-domain Character Distance for Robust Text Recognition}.
\newblock \emph{IJCV}, 132(2): 300--318.

\bibitem[{Zhu et~al.(2021)Zhu, Gao, Wang, and Wu}]{ssbaseline}
Zhu, Q.; Gao, C.; Wang, P.; and Wu, Q. 2021.
\newblock {Simple is Not Easy: A Simple Strong Baseline for TextVQA and TextCaps}.
\newblock In \emph{AAAI}, volume~35, 3608--3615.

\end{thebibliography}

\end{document}